# Cross-Language Personal Name Mapping

**Abstract**

Name matching between multiple natural languages is an important step in cross-enterprise integration applications and data mining. It is difficult to decide whether or not two syntactic values (names) from two heterogeneous data sources are alternative designation of the same semantic entity (person), this process becomes more difficult with Arabic language due to several factors including spelling and pronunciation variation, dialects and special vowel and consonant distinction and other linguistic characteristics. This paper proposes a new framework for name matching between the Arabic language and other languages. The framework uses a dictionary based on a new proposed version of the Soundex algorithm to encapsulate the recognition of special features of Arabic names. The framework proposes a new proximity matching algorithm to suit the high importance of order sensitivity in Arabic name matching. New performance evaluation metrics are proposed as well. The framework is implemented and verified empirically in several case studies demonstrating substantial improvements compared to other well-known techniques found in literature.

# 1. Introduction

Sharing data between organizations has growing importance in many data mining projects. Data from various heterogeneous sources often has to be linked and aggregated in order to improve data quality. Also, linked data can contain information that is not available otherwise and used to enrich data that is used for pattern detection in data mining systems. In the higher education sector, this includes the linking of scholar data from citation databases or eLearning initiatives participants' database to the management information system. In the health sector, information retrieved from linked data is used to improve health policies with census data. Businesses link their data sets to compile mailing lists. Data linkage is also used in crime and terror detection.

Often, in the real world, entities have two or more representations in databases. Usually, the databases contain information about people like names, emails, addresses and more. While there is only one correct spelling for many words, there are several valid spelling variations for persons' names. With the fast growth of enterprise integration application, the problem of joining data that contains person names from different applications is recognized. Person name records contain errors that make record matching a difficult task. Errors are introduced as the result of transcription errors, incomplete information, lack of standard formats, or any combination of these factors.

The name matching problem becomes more difficult if person names are represented in one language in one application and in another language in other application. The new problem of cross language entity linking is a new one and defined in 2011 by [1]. A new test collection is used to evaluate cross-language entity linking performance in twenty-one languages including Arabic and English [1]. This leads to the problem of Cross Language Personal Name Mapping (CLPNM). CLPNM supports the process of finding related records written in different languages using an automated system. This concept is used further in cross language information retrieval [2].

**Motivations**

Data sources containing data in non-Latin languages like Arabic language cannot be linked using the available record linking tools. Also, there are no available solutions to link person names between Arabic and English. There is a need to automatically build names dictionaries to support these solutions. Arabic based similarity functions are needed to facilitate record linking. Also there is a need to have a simple agreed upon quality measure for the record linking process between two languages.

**Contribution**

The contribution of this paper is significant for a number of reasons. It proposes a dictionary based framework for cross language name mapping along with its implementation. Furthermore, the paper proposes a strategy to automatically build names dictionary using an enhanced Arabic Soundex algorithm. A new enhanced weighted atomic token function is suggested. New evaluation metrics are proposed and used to evaluate the results. Results show the effectiveness of the proposed framework, functions and algorithms.

**Paper Outline**

The remainder of this paper is organized as follows: Section 2 presents the basic record linking background. Section 3 presents the related work. The proposed framework and architecture are shown in section 4. In section 5, the results of applying the proposed record linking methodology on several experiments are presented with thorough analysis and discussions. Finally, section 6 concludes the paper.

## 2. Background

An increasingly important task in the data preprocessing step of many data mining projects is the linking of two datasets to find records that are related to the same entity. This step is often required as information from multiple sources needs to be integrated, combined or linked in order to allow more detailed data analysis or mining. The aim of such linkages is to match and aggregate all records related to the same entity (e.g. people, organizations or objects) [3],[4]. Several frameworks were proposed to achieve these tasks [5].

Data linkage techniques have traditionally been used in the health sector and in statistics for linking census and related data. Computer-assisted data linkage goes back as far as the 1950s. Fellegi and Sunter put the mathematical foundation of probabilistic data linkage in 1969 [6]. This could be extended also to many applications in the higher education sector to support integration between e-learning, mobile quizzes, digital libraries and management information systems [7-10].

What makes name matching a problem is the fact that real-world data quality is low in most cases. Name matching can be viewed as related to the similarity search (wild card search). This paper focuses on person entities, when the identifier is the person name.

### 2.1. The Data Linkage Process

For two data sources A and B, the set of the ordered record pair resulting from cross joining the two data source AxB is the union of three disjoint sets, M, U and P [11]. The first set M is the matched set where the two records from A and B are equivalent. The second set U is the unmatched set where the two records from A and B are not equivalent. The third set P is the possible matched set. In the case that a record pair is assigned to P, a domain expert should manually examine this pair. It may be possible for the expert to judge if the record can be moved to either M or U.

As most real-world data collections contain noisy, incomplete and incorrectly formatted information, data cleaning and standardization are important preprocessing steps for successful data linkage. Data may be recorded or captured in various, possibly obsolete formats and data items may be missing, out of date, or contain errors. The cleaning and standardization of names and addresses is especially important to make sure that no misleading or redundant information is introduced (e.g. duplicate records). Names are often reported differently by the same person depending upon the organization they are in contact with, resulting in missing middle names or even swapped name parts.

There are many applications of computer-based name matching algorithms including record linkage and database searching that solve variations in spelling, caused for example by transcription errors. The success of such algorithms is measured by the degree to which they can overcome discrepancies in the spelling of names, in some cases it is not easy to determine whether a name variation is a different spelling of the same name or a different name altogether. Spelling variations can include misplaced letters due to typographical errors, substituted letters (as in Mohamed and Mohamad), additional letters (such as Mohamadi), or omissions (as with Mohamed and Mohammed). This type of variations in writing names doesn't affect the phonetic structure of the name. These variations mainly arise from misreading or mishearing, by either a human or an automated device. Phonetic variations appear when the phonemes of the name are modified, e.g. through mishearing, the structure of the name is substantially altered. Alternate first names problem appear in western languages when a person changes his name during the course of his life, usually when his marital state changes from single to married and vice versa. Double names occur in some cases where names are composed of two syllable but both are not always shown. For example, a double name such as Philips-Martin may be given in full, as Philips or as Martin. Double first names: Although not common in the English language, when considering Arabic, for example, names such as Abdel-Hamid may be given in full, or as Abdel and/or Hamid.

4Some of the difficulties associated with Arabic language person names were addressed in [12]. These difficulties include typographic variations like those shown in table 1 below.

Table 1: Typographic variants of Arabic names and English transliterations

| Typographic variations | Arabic Examples | English Transliterated Equivalents |
|---|---|---|
| The drop of hamza initially, medially, and finally | أحمد x احمد<br>إيمان x ايمان<br>فؤاد x فواد | Ahmed<br>Eman or Iman<br>Fuad or Foad |
| Two dots inserted on aleph maqsura, and two dots removed from yaa | منى x مني<br>هاني x هانى | Mona<br>Hani or Hany |
| Dropping the madda from the aleph | آلاء x الاء | Alaa |
| Hamza insertion below vs. above aleph | أحمد x إحمد | Ahmed |
| Two dots inserted on final haa, and two dots removed from taa marbouta | فاطمة x فاطمه | Fatma |
| Diacritics: partial, full, or none. Diacritics are removed. | عُمر x عمر | Omar |
| Typing hamza followed by aleph maqsura separately vs. together. | هانئ x هانىء | Hany |

Two types of data heterogeneity should be distinguished. They are structural and lexical. Structural heterogeneity appears when the schema and fields of the two databases are different. The Lexical heterogeneity occurs when the tuples have identically structured fields across databases, but the data use different representations to refer to the same real-world object. This includes the complication of joining name keys from two or more tables which are not in the same language. For example, all the following names are equivalent: Abd El Fatah, Abd-El-Fatah, عبد الفتاح. The use of dictionaries can help overcoming these difficulties.

There are several phonetic name matching algorithms including the popular Russell Soundex [13, 14] and Metaphone algorithms that are designed for use with English names. The ambiguity of the Metaphone algorithm in some words limited its use. The Henry Code is adapted for the French language while the Daitch-Mokotoff Coding method is adapted for Slavic and German spellings of Jewish names.

The Russell Soundex Code algorithm [13, 14] is designed primarily for use with English names and is a phonetically based name matching method. The algorithm converts each name to a four character code, which can be used to identify equivalent names, and is structured as follows:

1. Retain the first letter of the name, and drop all occurrences of a, e, h, i, o, u, w, y in other positions.
2. Assign the following numbers to the remaining letters after the first. The code values are shown in next table for both English and Arabic.
3. If two or more letters with the same code were adjacent in the original name (before step 1), omit all but the first.
4. Convert to the form 'letter, digit, digit, digit' by adding trailing zeros (if there are less than three digits), or by dropping rightmost digits if there are more than three.

The Arabic version of the Soundex algorithm is found in [15] and modified in [16]. Its approach is to use Soundex of conflating similar sounding consonants. Table 2 shows the character codes of English Soundex and Arabic Soundex.

TABLE 2: SOUNDEX AND ARABIC SOUNDEX CODING
This table shows the coding of characters in Soundex algorithm

| English Character | Code Value | Arabic Character |
|---|---|---|
| A, E, H, I, O, U, W and Y | 0 | ا، أ، إ، آ، ح، ع، هـ، و، ي |
| B, P, F, and V | 1 | ب،ف |
| C, S, K, G, J, Q, X, Z | 2 | خ، ج، ز، س، ش، ص، غ، ق، ك |
| D, T | 3 | ت،ث، د،ذ، ض، ط، ظ |
| L | 4 | ل |
| M, N | 5 | م،ن |
| R | 6 | ر |

**Edit (Levenshtein) Distance and Atomic String**

Levenshtein distance [17] is a character-based similarity measure of the similarity between two strings, which we will refer to as the source string (*s*) and the target string (*t*). The distance is the number of deletions, insertions, or substitutions required to transform *s* into *t*. the greater the Levenshtein distance, the more different the strings are. For example, if we have the strings "Hamed" and "Mohamed", the Levenshtein distance is 2.

Monge and Elkan [18] proposed a token-based metrics for matching text fields based on atomic strings. An atomic string is a sequence of alphanumeric characters delimited by punctuation characters. Two atomic strings match if they are equal or if one is the prefix of the other.

## Quality of the record linking techniques

The quality of record linking techniques can be measured using the confusion matrix as discussed in [11] that compares actual matched (M) and non-matched (U) records (using domain expert) to the machine matched (M') and non-matched records (U'). Well known measures include true positives (TP), true negatives (TN), false negatives (FN), and false positives (FP).

The measurement of accuracy, precision and recall are usually expressed as a percentage or proportion as follows [11]:

$$\text{Accuracy} = (TP + TN) / (TP + FP + TN + FN).$$
$$\text{Precision} = TP / (TP + FP)$$
$$\text{Recall} = TP / (TP + FN)$$

Because the number of negatives TN is very large compared to the number of records in the comparison space, it is widely accepted that quality measures that depend on TN (like accuracy) will not be very useful because TN will dominate the formula [11]. In this paper, new metrics will be proposed to solve these problems.

## 3. Related Work

In order to perform a join between two relations without a common key, it is needed to determine whether two specific tuples, i.e. field values are equivalent [2]. Entity matching frameworks provide several methods as well as their combination to effectively solve different matching tasks. In [3], eleven proposed frameworks for entity matching are compared and analyzed. The study stressed the diversity of requirements needed in such frameworks including high effectiveness, efficiency, generality and low manual effort. In [4], the characteristics of personal names and sources of variations and errors are discussed. Experimental comparisons of several large name data sets indicate that there is no clear best technique to solve the name mapping problem [4]. The de-duplication is a similar problem to record linking when the source and destination are the same. A thorough analysis of the literature on duplicate record detection is presented in [5] where similarity metrics and several duplicate detection algorithms are discussed.

There are three basic types of record linkage strategies: deterministic, probabilistic and modern [19]. The deterministic approach can be applied if high quality precise unique entity identifiers are available in all the data sets to be linked. At this level, the problem of linking at the entity level becomes trivial: a simple database join is all that is required. However, in most cases no unique keys are shared by all the data sets, and more sophisticated linkage techniques need to be applied. In probabilistic linkage, the process is based on the equivalence of some existing common attributes in the data sets. The probabilistic approach is found to be more reliable, consistent and provides more cost effective linkage results. The modern approaches include approximate string comparisons and the application of the expectation maximization (EM) algorithm [3]. Many Other techniques are explored including machine learning, information retrieval and database research. Some frameworks utilize training data to semi-automatically find an entity matching strategy to solve a given match task. The quality of the computer linked records process is found to be higher than the manually linked record (done by hands of humans) [20].

The record linking problem is extended in several ways. The problem of carrying out the linkage computation without full data exchange between the two data sources has been called private record linkage and discussed in [21]. The problem to identify persons from evidence is the primary goal of forensic analysis [22]. Machine translation in query translation based cross language information access is studied in [23]. Speech-to-speech machine translation is another extension that can be achieved using grapheme-to-phoneme conversion [24]. The problem is extended in another way to match duplicate videos and other multimedia files including image and audio files. This increases the need to have high performance record linking [25].

In cross language information retrieval, it is found that a combination of static translation resources plus transliteration provides a successful solution [20]. As normal bilingual dictionaries cannot be used for person names, these person names should be transliterated because they are considered out of vocabulary (OOV) words. A simple statistical technique to train English to Arabic transliteration model from pairs of names is presented in [21]. Additional information and relations about the entities being matched could be extracted from the web to enhance the quality of linking [22].

Many algorithms are proposed that depends on machine learning approaches [10], [20], [28], [29], [30], [31]. One of the major challenges when linking large databases is the efficient and accurate classification of record pairs into matches and non-matches. Traditional classification is based on manually-set thresholds or on statistical procedures. More recently developed classification methods are based on supervised learning techniques. They therefore require training data, which is often not available in real world situations or has to be prepared manually by a time-consuming process. Several unsupervised record pair classification are presented in [26, 27]. The first is based on a nearest-neighbour classifier, while the second improves a Support Vector Machine (SVM) classifier by iteratively adding more examples into the training sets.

The problem of record matching in the Web database scenario is addressed in [28]. Several techniques to cope with the problem of string matching that allows errors are presented in [29]. Many fast growing areas such as information retrieval and computational biology depend on these techniques.

Machine transliteration techniques are discussed in [2, 30, 31] for Arabic and Japanese languages. In [3], finite state machines are used with training of spelling-based model. Statistical methods are used for automatically learning a transliteration model from samples of name pairs in two languages in [32, 33]. Machine translation could be extended later from text to speech as found in [24]. Co-training algorithms with unlabeled English-Chinese and English-Arabic bilingual text is used in [34]. A system for Cross Linguistic Name Matching in English and Arabic is implemented in [1, 35]. The system augmented the classic Levenshtein edit-distance algorithm with character equivalency classes.

Several tools are used for matching names in Latin languages including Febrl, TAILOR and BigMatch. They use different techniques to identify any identical entities from two or more data sources in the same language. Febrl [36] provides data structures that allow efficient handling of very large data sets. The results of a survey of Febrl users is discussed in [37]. Febrl includes a new probabilistic approach for improved data cleaning and standardization that support parallelization [38]. TAILOR [39] is a flexible record matching toolbox which allows the users to apply different duplicate detection methods on the data sets. BigMatch Is the duplicate detection program used by the US Census Bureau [40]. If the sizes of the datasets are large, online record linking can be used [41]. Tools that support Arabic language used some hybrid algorithms for name matching and duplicate record detection [42], [43].

However, the aforementioned tools do not support Arabic language based names and/or transliterations because most of them do not support the Unicode system. Also they are not aware with the structure of Arabic names and their characteristics.

## 4. Methodology

In order to link names between different languages, like English and Arabic, a dictionary is built once using the data of the source and destination database. Then the dictionary is used as an interface for transliteration and other matching exercises. A search process will be executed to find matched records. In the following subsections, the proposed framework will be presented, the details of the preprocessing stage will be described, the dictionary building alternatives will be suggested, the searching process will be described in details and finally the new proposed quality metrics will be presented.

### 4.1. The proposed framework

The architecture of the proposed system is shown in figure 1. Figure1 outlines the general steps involved in the linking of two databases in different languages. A necessary step in any record linkage solution is the data cleaning and standardization as real-world databases contain always dirty and noisy, incomplete and incorrectly formatted information. The main task of data cleaning and standardization is the conversion of the raw input data into well defined, consistent forms, as well as the resolution of inconsistencies in the way information is represented.

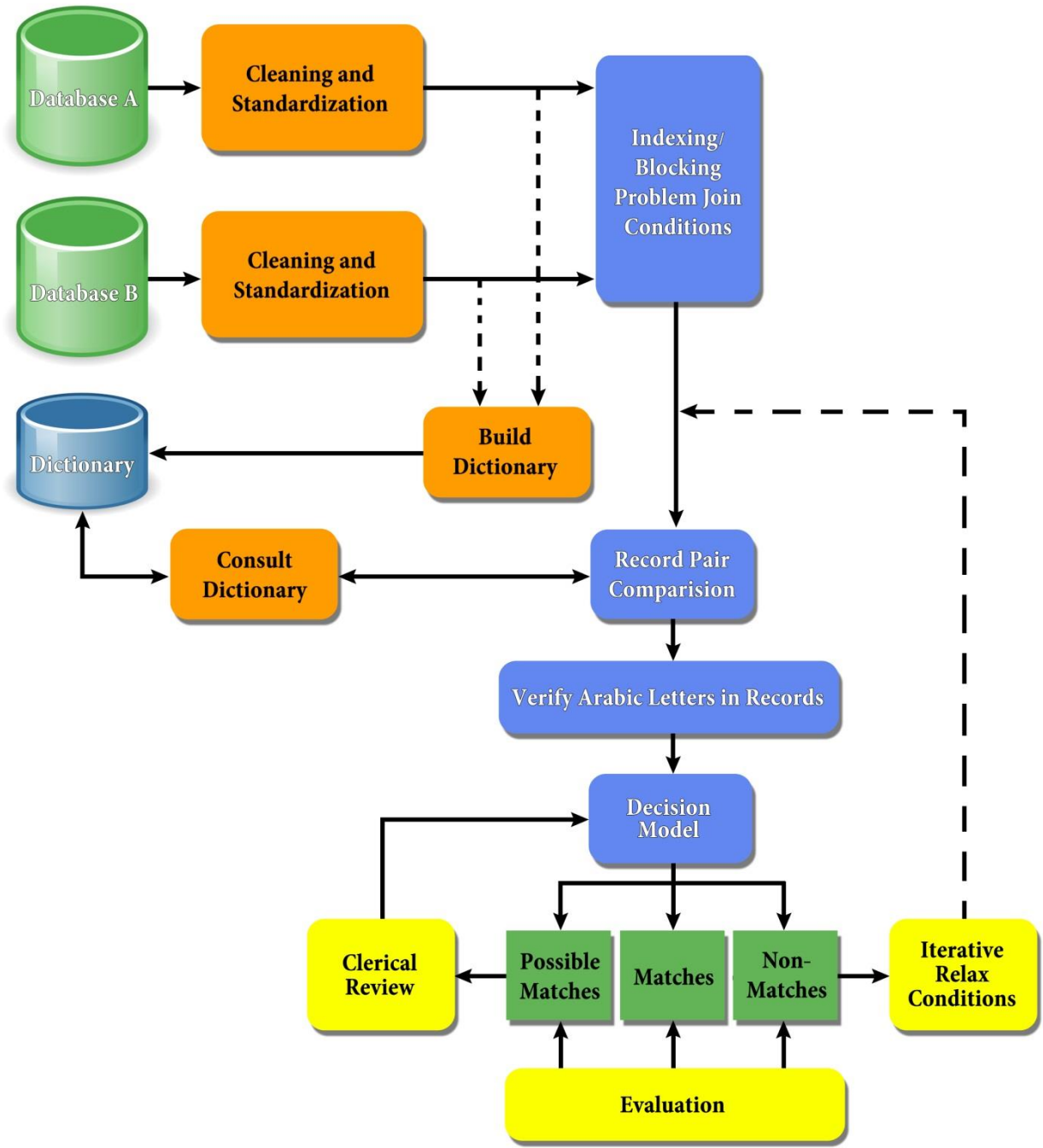

Figure 1: The proposed framework

The second step ('Indexing/Blocking') applies the problem domain join conditions to eliminate clear unmatched records and generates then pairs of candidate records. These records are compared in detail using approximate string comparisons, which take (typographical) variations into account. Then, the results are verified and a decision model is used to classify the compared candidate record pairs into matches, non-matches, and possible matches.

The non-matched records are passed to the iterative relax conditions process in order to reduce the search conditions in hope of getting matched records. Clerical review process is used to manually assess the possible matched pairs and classify them into matches or non-matches. Clerical review is the process of human oversight to decide the final linkage status of possible matched pairs. The person undertaking clerical review usually has access to additional data which enable them to resolve the linkage status, or applying human intuition or common sense to decision based on available data. Measuring and evaluating the quality and complexity of a record linkage project is a final step in the record linkage process. Many new quality metrics are suggested and proposed in this paper to help the evaluation of the efficiency and quality of record matching process between dual languages.

*4.2. Pre-Processing: Cleaning and Standardization*

The matching process is preceded by a data preparation stage. This is shown in figure 1 with the orange processes. During the data preparation stage, data is unified, normalized and standardized. These steps improve the quality of the in-flow data and make the data comparable and more usable. It simplifies the recognition of Arabic name and their typographic variants. The pre-processing step is done on several levels including character level normalization, splitting and parsing and combined names canonical format generation.

English Names character normalization includes removing separators like additional spaces, hyphens, underscores, commas, dots and slashes from the full names. For example, the following two names are equivalent (Mohamed, Abd El-Fattah and Mohamed Abd El Fattah). This includes also converting every uppercase letter to a lowercase letter. Standardizing Arabic names through character normalization is more difficult and involves several steps. The first step is replacing each character appearing in the left column with the unified character shown in the second column of the table below.

Table 3: Character Normalization

| Equivalent characters | Unified character |
|---|---|
| أ ، إ ، آ ، ا | ا |
| ى | ي |
| ة | ه |
| ؤ | و |

Names standardization refers to the process of standardizing the Arabic names and their English transliteration into a specific uniform content format representation. This is done in several steps as follows:

*4.2.1. Splitting and Reordering*

After data Pre-Processing take place, the English and Arabic full names are split into separate names as shown below in Figure 2.

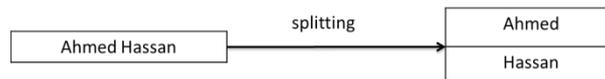

Figure 2: Persons' Full Names Splitting

In several applications, English names are represented in a format when last names appear first. In Arabic language, first names appear first. Changing the order of the English names to match the Arabic names is an important step to align the names.

*4.2.2. Name Parsing and unification (canonical form generation)*

The majority of Arabic names and their transliterations consist of one syllable. However, names can be composed of more than one syllable. These names have either prefixes like (Abdel Rahman, Abd El Aziz, Abou El Hassan, Abou El Magd) or postfixes like (Seif El Din, Hossam El Din). A list of used spelling variants of Arabic names prefixes in Arabic language, transliterated into English language, includes: Abd, Abd Al, Abd El, Abdel, Abdol, Abdul, Abo, Abo El, Aboel, Abou, Abou El, Abu, Al, El. Postfixes include: El Din, El Deen, Allah and others. Postfixes can be treated easily with the iterative relax conditions process, which will be described later. However, names with prefixes should be parsed to identify them and put them in a canonical form. For example, a full name like "Abdel Rahman Mohamad" is split as shown in the left of the next figure, appears as if it consists of three names. The name parsing process uses the pre-stored prefixes table to reorganize "Abdel Rahman" as a single name as shown in Figure 3.

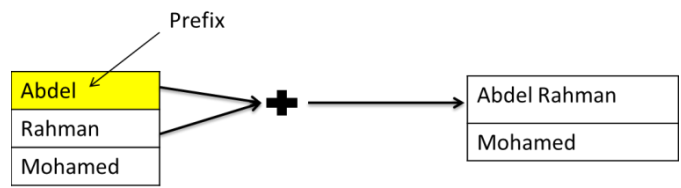

Figure 3: Parsing the transliterated Arabic compound names

The last step here is the unifying process which unifies the variants of "Abdel Rahman" including "Abd El Rahman", "Abdul Rahman", "Abd Al Rahman" to a single unified canonical form.

*4.3. Dictionary building*

After cleaning and standardizing the two datasets, a dictionary should be built before starting the search process. This dictionary will contain a record for each Arabic Character and the corresponding English Equivalents. It will contain also the list of all Arabic names and their English transliterated equivalent. It is important here to notice that one English letter can correspond to several Arabic letters. For example, the "E" letter can correspond to ("ا", "ي", "ع"). Therefore, it is important to have all the possible transliterations of a name and add them to the dictionary. There are three possibilities to build the dictionary that will be discussed here. An experiment is designed to compare the accuracy of these dictionaries. The results of applying this experiment will be shown in the results section. The different possibilities are:

*4.3.1. Source Dataset extracted dictionary*

Usually, the dataset containing full Arabic names has a field that contains the English transliteration of these names. The data in these fields are usually entered by data entry operators. Although this data can be used together to form the dictionary, the accuracy and performance of such a dictionary will be very low because a lot of wrong data entries will be found.

*4.3.2. Phonetic based dictionaries*

Because names from different language are nearly pronounced with the same pronunciation and shared the same phonetic attributes, phonetic algorithms described in previous sections could be used to build dictionaries from the first dataset and second dataset for record pairs that have the same phonetic code. This code can be used as a join condition. These algorithms and the techniques used to generate the dictionaries will be presented here.

*4.3.2.1. Soundex generation and dictionary*

Soundex technique matches similar phonetic variants of names. Because names from different language are nearly pronounced with the same pronunciation, this technique is used with the Arabic Soundex algorithm to create the same code of Arabic names as shown in Figure 4.

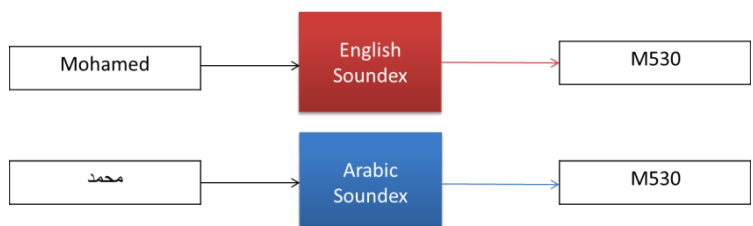

Figure 4: Applying Soundex for Arabic names and their English transliterated equivalents

After creating the code for every English and Arabic name in the two datasets, a join operation can be used to match the code of similar English/Arabic names to create a dictionary as shown in table 5.

Table 4: English/Arabic names dictionary using Soundex algorithm

| Arabicname | Englishname | ArabicSoundex | EnglishSoundex |
|---|---|---|---|
| بكير | Bakir | B260 | B260 |
| بلال | Belal | B440 | B440 |
| بلتاجي | Beltagi | B432 | B432 |

*4.3.2.2. Enhanced Arabic Combined Soundex Algorithm and Dictionary*

Although applying the Soundex and Arabic Soundex works well with Arabic names with one syllable, it does not work with Arabic names that are composed of more than one word like (Abdel Aziz, Aboul Hassan, Essam El Din). For example Abdel Aziz will have the same Arabic Soundex of (Abdel Rahman) because the Soundex will code those first four consonant letters only so they both will have the same code (A134). This is shown in Figure 5.

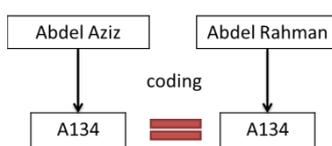

Figure 5: Effect of using Soundex Algorithm on Arabic combined name

A new technique was proposed to identify the names that start with prefixes called "Arabic Combined Soundex". Arabic Combined Soundex algorithm splits the prefix from the second syllable and creates a code for both of them. Both codes (4 characters) are then concatenated together to form a combined Soundex code with 8 characters. Using this technique help in distinguishing between "Abdel Aziz" and "Abdel Rahman" mentioned in the previous example, where "Abdel Aziz" will have the combined code of (A134A220) and "Abdel Rahman" will have the combined code (A134R550) as shown in Figure 6.

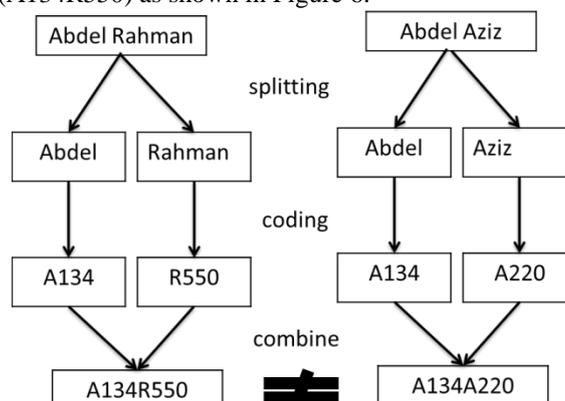

Figure 6: Using Arabic Combined Soundex to resolve combined names

*4.3.2.3. Other Considerations for Phonetic Algorithms in English transliterated Arabic Names*

It is found that some English transliterated Arabic names have different English Soundex and Arabic Combined Soundex. For example, "Ola" is the English transliterated version of the Arabic name "علا". However, they have different Soundex in English and Arabic. The reason is that the Soundex technique appends the first letter to the code and the letter ع could be translated in English to "A", "E" or "O" depending on the next Arabic character or diacritics. Therefore, the Soundex code of "Ola" will be (O400) and the three versions of Arabic Soundex should be tried (A400, E400, O400). The same problems appear for "Yousef", "Uossef" which has a Soundex code starting with "Y" and "U", respectively. However, their Arabic name (يوسف) has always a code starting with "Y". The previous examples show the need for some human work to complete the dictionary.

*4.3.3. User verified combined Soundex Dictionary*

This dictionary used the combined Soundex Dictionary, and then domain experts' work was done to verify the dictionary and add entries that increase its accuracy and performance. It is expected that the matching process that uses this dictionary will have higher degree of accuracy and performance compared to the aforementioned dictionaries.

*4.4. Searching*

The searching process is composed of several steps, shown in figure 1 with blue color. They are the indexing/blocking step, record pair comparison process, verification process, decision process and the iterative relax conditions process. They will be described in the next subsections.

*4.4.1. Indexing/Blocking*

Each record linkage problem is associated with some problem domain join conditions. These conditions identify which record pairs are possible candidates. Field matching is used to match equivalent or similar values in corresponding fields between the two data sets. These fields are usually additional fields, other than the name fields. Field matching is used also as a blocking scheme that minimizes the number of record pairs to be compared later. The indexing/blocking step increases the relevance of the subsequent search steps.

An example of this indexing/blocking is searching for death cases for citizens from certain governorate from the first dataset and citizens from the same governorate from the census data set. Here, the governorate is used as an indexing/blocking condition. Records that are not satisfying the condition will be assumed true negatives and not considered as matches or possible matches.

*4.4.2. Record Pair Comparison*

After generating the candidate record pairs in the indexing/blocking step, the record pair comparison step is executed. For each record from the first source dataset, a full name is extracted, reordered if it is in English, split into the *n* names constituting it and finally parsed. If the first source is in English, the dictionary is consulted and the corresponding Arabic names are retrieved. These *n* Arabic names are combined with the SQL '%' wildcard operator to generate search conditions. For example, if we are searching for the English full name 'Mohamed Abdel Fattah Salama', the generated SQL search condition will be "like "محمد%عبد الفتاح%سلامة%"". These search conditions are then applied to record pairs for comparison. Record pairs that satisfy search conditions are then passed to the decision model classifier. Source dataset records that have no corresponding destination records will be marked and passed to the iterative Relax Condition process. Record pairs that satisfy search condition are then passed to the decision model classifier.

*4.4.3. Verification Process*

When the machine searches for Arabic names by its English transliterations including initial letters, the results contains some false matched records as a result of the one to many mapping between English and Arabic letters.

The verification process uses the dictionary again in a reverse way to verify the results and enhance the quality of the matching process. For example, through standardization process, the Arabic characters 'أ، إ، ا' are replaced by 'ا' to prevent typographical errors. When the search process searches for a person name with initial character 'A', the dictionary retrieves all names started with ("ا") including "ايمان" although its transliteration is 'Eman' which does not start with character 'A'. The verification process uses the dictionary in a reverse way (from Arabic to English) and find that the name ("ايمان") is translated to ("Eman") and the name's first character (E) is not the same as the search condition with character 'A', and then this record is eliminated form the result set.

*4.4.4. Decision Model: Weighted Atomic Token*

When a record from the source dataset corresponds to only one record pair, this record pair is assumed a matched record. When several record pairs satisfy the conditions generated from a record from the source dataset, a technique should be used to sort these record pairs and put the most probable records in the matched set. Records with less probability value will be put in the possible match dataset. If it is possible to put some record pairs at this stage into the unmatched dataset, this will reduce the amount of work done in the clerical review process.

The accuracy of the results can be enhanced by decreasing the result record pairs that corresponds to each name in the first source dataset. Weighted Atomic token is used to determine the best matches from all the results. The proposed weighted atomic token calculates a percentage of similarity between two strings as shown in the following example and multiplies it with the distance between the similar words. Although the traditional atomic token does not take into consideration the order of the two strings, the weighted atomic token takes it into account. In the following section, a detailed description of weighted atomic token is presented in the following table.

Table 5: Example of the Weighted Atomic Token algorithm

| | | 4 | | 3 | | 2 | | 1 | | | |
|---|---|---|---|---|---|---|---|---|---|---|---|
| | | سلامة | | علي | | ف | | أحمد | | | |
| 1 | أحمد | 0 | 3 | 0 | 2 | 0 | 1 | 1 | 0 | | |
| 2 | فاروق | 0 | 2 | 0 | 1 | 0.2 | 0 | 0 | 1 | | |
| 3 | سلامة | 1 | 1 | 0 | 0 | 0 | 1 | 0 | 0 | | |

In the previous table, the numbers shown in red represents the similarity between the two names in the same row and column. Such similarity is calculated as the number of similar characters divided by the number of characters of the longer string. The numbers shown in blue represent the distance between the orders of each substring.

Assume we have two string *s1*, *s2* where string *s1* contains *n1* substrings and *s2* contains *n2* substrings and where *n2>=n1*.

The atomic token is defined by the following equation [18].

$$\text{AT} = (\sum_{k=1}^{n1} Sim(substr(s1,k), substr(s2,k)))/n2$$

Where
    AT is the Atomic token value.
    Sim (a, b) = number of similar consecutive characters of (a, b) / maximum (length(a), length(b)).
    Substr (s, k) is the k$^{th}$ substring of the string s.
Then the Atomic Token value for the example shown above = (1 + 0.2 + 0 + 1) / 4 = 0.55.

The weighted atomic token is defined by

$$\text{WAT} = \sum_{i=1}^{n2} \sum_{k=1}^{n1} (1 - dist(i,k)/n2) * Sim(substr(s1,k), substr(s2,i)))/n1$$

Where
    WAT is the weighted atomic token

dist (i, k) = abs (i - k)

In the previous example, the Weighted Atomic Token = (1*1 + 0.2*1 +1*.75) / 3 = 0.65.

We should also mention that the splitting of substrings here satisfy the aforementioned considerations of combined names in Arabic language.

*4.4.5. Iterative Relax Condition*

Relax condition is a process that reduces the number of search conditions in order to get more records in the results dataset. If a full name containing *n* names (first name, last name and *n-2* names in between) is searched and no results is retrieved, this process searches for *n-1* names, including the first and the last names, eliminating one of the names in the middle. This process is useful when the full name to be searched contains more details than the full names found in the destination database. This process is iterative and tries to relax more conditions if no results are retrieved by eliminating one condition only. The process is then used by eliminating the last name and searching for the first n-1 names. At the last step, the process uses Levenshtein distance to solve for any typographical errors in the first name.

*4.5. Proposed Quality Evaluation of Cross Languages Record Linkage*

It is clear that the confusion matrix (described in table 3 in the background section of this paper) has dropped the cases of possible matches. Because the number of possible matches in record linking is very large, especially in the case of the cross languages record linking, we developed some new metrics that take this consideration into account.

Because the person names record linkage problem is usually a one to one relationship. For a record a in the source dataset *A*, the machine can identify that the record have one exact match in the Match set (M'). So this case is denoted by 1..1. If the record a has many records in the non-matched set (U) and no records at all in the matched and possible matched sets, this record is not found and this case will be denoted by 1..0. If the record has *q* several equivalent records in P', this means that clerical reviewer should decide which one of these *q* possibilities is the most realistic and this case is denoted by 1..*q*. It is clear that smaller *q* is better for the clerical reviewer. This means that the machine accuracy calculation should take into consideration how useful the machine was to simplify the work for the domain expert. We developed an algorithm to measure the matching accuracy of the results.

Assume that we have the following results presented in table 7 after being evaluated by the subject matter experts.

Table 6: Example of Matching Results after being evaluated by the subject matter experts

| Machine | Source Dataset A | Source Dataset B | Similarity | Evaluation | Other records found by the subject matter expert | Subject Matter Expert |
|---|---|---|---|---|---|---|
| 1..3 | A1 | B1 | 0.9 | Accepted | | 1..1 |
| 1..3 | A1 | B2 | 0.7 | Not Accepted | | 1..1 |
| 1..3 | A1 | B3 | 0.5 | Not Accepted | | 1..1 |
| 1..0 | A2 | Not Found | | Accepted | | 1..0 |
| 1..1 | A3 | B6 | 0.95 | Accepted | | 1..1 |
| 1..3 | A4 | B7 | 0.3 | Not Accepted | | 1..0 |
| 1..3 | A4 | B8 | 0.7 | Not Accepted | | 1..0 |
| 1..3 | A4 | B9 | 0.6 | Not Accepted | | 1..0 |
| 1..0 | A7 | Not Found | | Not Accepted | B32 | 1..1 |

| 1..2 | A8  | B10       | 0.7 | Accepted     |                    | 1..2 |
| 1..2 | A8  | B11       | 0.7 | Accepted     |                    | 1..2 |
| 1..1 | A7  | B23       | 0.2 | Not Accepted |                    | 1..0 |
| 1..0 | A8  | Not Found |     | Not Accepted | B22, B21, B20, B18 | 1..3 |
| 1..1 | A9  | B12       | 0.5 | Accepted     | B13, B14           | 1..3 |
| 1..2 | A10 | B15       |     | Accepted     | B17,B18            | 1..4 |
| 1..2 | A10 | B16       |     | Accepted     | B17,B18            | 1..4 |

The data is classified in the confusion matrix as follows.

|  |  |  | Machine | | | | | | | |
|---|---|---|---|---|---|---|---|---|---|---|
|  |  |  | U | M | P | | | | | |
|  |  |  | 1..0 | 1..1 | 1..2 | 1..3 | 1..4 | 1..5 | ... | $1..N_{M_{Max}}$ |
| Subject Matter Expert | U | 1..0 | TPP | FNP |  |  |  |  |  |  | ← Block D |
|  | M | 1..1 | FPP | VTNP |  |  |  |  |  |  | ← Block C |
|  | P | 1..2 |  |  |  |  |  |  |  |  |
|  |  | 1..3 |  |  |  |  |  |  |  |  |
|  |  | 1..4 |  |  |  |  |  |  |  |  |
|  |  | 1..5 |  |  |  |  |  |  |  |  |
|  |  | ... |  |  |  |  |  |  |  |  |
|  |  | ... |  |  |  |  |  |  |  |  |
|  |  | $1..N_{S_{Max}}$ |  |  |  |  |  |  |  |  |
|  |  |  | ↑ Block B | ↑ Block A |  |  |  |  |  |  |

*4.5.1. Primary Quality Metrics*

The number of the confusion matrix cells increases now from the four original cells representing true positives (TP), true negatives (TN), false positives (FP) and false negatives (FN) to a large number of cells. In our proposed evaluation metrics, the definitions of TP, FP and FN will remain the same as found in [11]. The first change in the proposed metrics is the introduction of Verified Relevant True Negatives cell (VRTN or simply VTN). They are records from the first dataset for which the machine failed to find corresponding records from the second dataset (Unmatched records). These records are verified also by the subject matter expert as unmatched. It is a measure of the machine efficiency to differentiate records. This measure has a real value and does not have the pitfalls of true negatives (TN) described in [11].

According to the confusion matrix, the diagonal elements show the region in which the machine and subject matter experts have agreed upon results. The off diagonal elements show the region in which the machine and subject matter experts have different results. The difference becomes more sever when the distance between the element and the diagonal increases. The efficiency is directly proportional to the sum of the diagonal elements of the confusion matrix and inversely proportional to the sum of the weighted off diagonal elements where the weight is proportional to the distance between each element and the diagonal. Therefore, the effectiveness is defined as

$$Effectiveness = \sum_{i=1}^{N_{Max}} N_{i,i} / (\sum_{i=1}^{N_{Max}} N_{i,i} + \sum_{i=1}^{N_{SMax}} \sum_{k=1}^{N_{MMax}} abs(k-i)N_{i,k})$$

If the confusion matrix is a **strictly diagonally dominant** matrix, this shows that most of the results obtained by the machine are accepted by the subject matter expert.

The light olive green horizontal block (Block C) represents records that are classified by the machine as one to many matches and the subject matter experts can identify a single correct match from several matched records detected by the machine. The usefulness of such records is inversely proportional to the number of suggestions. This block is a measure of the effectiveness of the machine to detect similarity and offer suggestions. So, it should be added to the true positives percentage to get the extended true positive accuracy percentage ETPAP defined here. However, the accuracy here should be inversely proportional to the number of suggestions

$$Extended\ True\ Positive\ Accuarcy\ Percentage\ (ETPAP) = (N_{C_{1..2}}/2 + N_{C_{1..3}}/3 + \ldots + N_{C_{1..k}}/k)/N$$

$$Extended\ True\ Positive\ Accuarcy\ Percentage\ (ETPAP) = \left(\sum_{k=2}^{N_{M_{max}}} \frac{N_{C_{1..k}}}{k}\right)/N$$

$$Overall\ True\ Positive\ Accuarcy\ Percentage\ (OTPAP) = TP + ETPAP = \left(\sum_{k=1}^{N_{C_{max}}} \frac{N_{C_{1..k}}}{k}\right)/N$$

*4.5.2. Secondary Quality Metrics*

This horizontal light olive green area (Block C) is also characterized by the Extended Multiple False Identification (EMFI). EMFI is a measure for the inefficiency of the machine to decide and specify the best true positive match from the different offered suggestions. If the number of destination records identified by the machine for one source record is *k*, only one of them is verified by the subject matter expert, presenting *1/k* accuracy, then the false identification value will be *(k-1)/k*. Then, the Extended Multiple False Identification (EMFI) will be defined as follows:

$$EMFI = ((N_{C_{1..2}}/2) + (2N_{C_{1..3}}/3) + \ldots + ((k-1)N_{C_{1..k}}/k))/N$$

$$EMFI = \left(\sum_{k=2}^{N_{C_{max}}} (k-1)N_{C_{1..k}}/k\right)/N$$

Where

$N_{C_{1..k}}$ is the number of source dataset records that have k corresponding records from the second dataset identified by the machine while they are identified with one only corresponding record by the subject matter expert.

$N$ is the number of records of the smaller dataset

$N_{M_{max}}$ is the maximum number of repetitions of records from the second source dataset identified by the machine and corresponds to a single record from the first source dataset.

Now, the proposed accuracy, proposed precision and proposed recall will be redefined according to the previously mentioned measurements.

The proposed accuracy is redefined as

$$Proposed\ Accuracy = \frac{TPP + VTNP + ETPAP}{TPP + VTNP + ETPAP + FPP + EMFI + FNP}$$

Because VTN is mentioned in the formula, the calculated accuracy values will be realistic and will not be too high (as was the case when TN appears).

Proposed Precision is redefined as

$$Proposed\ Precision = \frac{TPP + ETPAP}{TPP + ETPAP + FPP + EMFI + FNP}$$

Proposed Recall is redefined as

$$Proposed\ Recall = \frac{TPP + ETPAP}{TPP + ETPAP + FPP + EMFI + FNP}$$

- The white horizontal block (Block D) represents records that are classified by the subject matter expert as non-match although several matched records are detected by the machine (extended multiple false positive, EMFP). This is a measure of the inefficiency of the machine to identify unmatched record.

$$EMFP = (2 * N_{D_{1..2}} + 3 * N_{D_{1..3}} + ... + k * N_{D_{1..k}})/(N_D)$$

$$EMFP = \left(\sum_{k=2}^{N_{M_{max}}} k * N_{D_{1..k}}\right) / (N_D)$$

Where

$N_{D_{1..k}}$ is the number of source dataset records that have k corresponding records from the second dataset identified by the machine while they are not identified to any corresponding record by the subject matter expert.

$N_D$ is number of records of the smaller source dataset in the D region

- The vertical light green olive block (Block A) represents records that are classified by the machine as one to one match and more matched records are detected by the subject matter experts (extended multiple true possible positives, EMTPP). This is a measure of the inefficiency of the machine to get all possible matches (lack of generality). If one record from the source dataset has four corresponding records identified by the domain experts, but the machine finds only one, this means that the machine has inefficiency that is proportional to 3 records out of four.

$$EMTPP = ((N_{A_{1..2}}/1) + (2N_{A_{1..3}}/3) + (3N_{A_{1..4}}/4) + ... + ((k-1)N_{A_{1..k}}/k))/ N_A$$

$$EMTPP = 1/N_A (\sum_{k=2}^{N_{A_{max}}} (k-1)N_{A_{1..k}}/k)$$

Where

$N_{A_{1..k}}$ is the number of source dataset records that have one exact match record from the second dataset identified by the machine and k records from the second dataset identified by the subject matter expert.

$N_A$ is total number of records of the first dataset in the A region

$N_{S_{max}}$ is the maximum number of repetitions of records from the second source dataset identified by the subject matter expert and corresponds to a single record from the first source dataset.

- The vertical white block (Block B) represents records that are classified by the machine as non-match and several matched records are detected by the subject matter experts. The Extended Multiple False Negative (EMFN) is a measure of the inefficiency of the machine to get all possible matches.

$$EMFN = ((2N_{B_{1..2}}) + (3N_{B_{1..3}}) + ... + (kN_{B_{1..k}}))/ N_B$$

$$EMFN = (\sum_{k=2}^{N_{B_{max}}} kN_{B_{1..k}})/(N_B)$$

Where

$N_{B_{1..k}}$ is the number of source dataset records that have 0 records from the second dataset identified by the machine and k records identified by the subject matter expert.

$N_B$ is number of records of the source dataset in the B region

$N_{S_{max}}$ is the maximum number of repetitions of records from the second source dataset identified by the subject matter expert and corresponds to a single record from the first source dataset.

The EMFN and EMFP are considered zeros when regions B and D respectively are empty. If the regions B and D are not empty, the EMFN and EMFP are always greater than 1 because number in destination records is more than number of source records in these regions.

## 5. Experimental Results and Discussions

The proposed framework is tested and verified with several experiments using data from the Higher Education sector in Egypt. In the following subsection, the details of the used data sources are discussed. Then, the experiments, results and discussion will be presented.

### 5.1. Data Sources

There are four used data sources to test the effectiveness of the proposed framework.

#### 5.1.1. Universities Management Information System (UMIS) Database

UMIS databases of the Egyptian universities have the same scheme and tables that contain totally about 80,000 staff members' data. The most important fields are shown in the table below:

Table 7: Sample Records from the Universities Management Information System (UMIS) Database

| UniID | FULL_NAME_AR | NATIONALNUM | STFDOB | EMP_DATE |
|---|---|---|---|---|
| 2 | سالي صلاح عنتر قاسم | 27554148800608 | 7/18/1982 | 10/31/2000 |
| 2 | سالي عبد المعطي | 27608555900303 | 4/14/1979 | 11/22/2001 |
| 2 | سامح سعد | 28210555900157 | 10/12/1982 | 9/26/2006 |

#### 5.1.2. Egyptian Universities Promotion Committees (EUPC) and eLearning Data

The Arabic data of several hundred faculty staff members who participated in the eLearning initiative is stored in a database. This database includes: Person Name, College, and University. Also, the Arabic data of the senior faculty staff members are stored in a database that contains about 9,000 records. This database contains similar data to the data found in universities management information system (UMIS) but with more details.

#### 5.1.3. Digital Library

The list of outstanding authors prepared by the Digital library unit is used for English/Arabic Matching. A list of 4000 authors contains the English transliterated names and other basic data as shown in table (4) below:

Table 8: Sample Records from the Digital Library Database

| UniID | Author | Faculty |
|---|---|---|
| 6 | Wadie, Bassem S | Medicine |
| 8 | Abdul-Kader, A. M | Science |
| 6 | El-Baz, Mahmoud Abdo | Engineering |

### 5.2. Experiments, Results and Discussion

Three experiments were performed on the data sources mentioned above. In the next subsections, each experiment is discussed with its results and analysis. The objective of linking the aforementioned sources is to study the characteristics of academic excellence in higher education and scientific research and to know the scientific fields subject to the brain drain problem.

*5.2.1. Experiment 1: Comparing the proposed framework to Levenshtein Edit distance/ indexing*

The first experiment aims to match **Arabic data** representing 335 faculty members who participated in the eLearning initiative in Egypt with the official **Arabic data** of all faculty members that is stored in the universities management information system (UMIS) database. In this experiment, these results are compared with the **Levenshtein Edit distance/ indexing** approach using the confusion matrix approach.

Table 9: Comparing Results of the proposed framework and the Levenshtein Edit Distance Approach

|  | Proposed Framework | **Levenshtein Edit distance/ indexing** approach |
|---|---|---|
| True Positives Percent | 89.55% | 70.15% |
| False Positives Percent | 1.49% | 28.96% |
| Verified True Negatives Percent | 5.37% | 0.00% |
| False Negatives Percent | 3.58% | 0.90% |
| Original Precision | 98.36% | 70.78% |
| Original Recall | 96.15% | 98.74% |
| Overall True Positive Accuracy Percentage | 91% | Not Applicable |
| Efficiency |  |  |

It is clear that the proposed framework has higher value of true positives and verified true negatives and less number of false positives compared to the Levenshtein distance framework. These results show the efficiency of the proposed framework. The proposed framework has the advantage of suggesting 2 or more alternatives if the best match of the person name is not decided. Therefore, the number of false positive is very small compared to Levenshtein. However this advantage causes a disadvantage of higher rate of false negatives and less recall value because the algorithm tolerance to typographical errors is large. This could be corrected by increasing the similarity/matching threshold value.

It is worth mentioning here that the use of the indexing/blocking as a problem domain join conditions enhanced the performance and narrowed the block to be searched. Therefore, less relevant record pairs are suggested as candidates of matching. The maximum number of occurrences in the proposed framework found was 4. This increased the overall true positive accuracy percentage. Also, the Relax condition process played a major role when the source and destination person names are not equal in length. It is found that Levenshtein failed in 97 cases due to the similarity between Arabic Person names of "محمد" and "أحمد" which has just one character difference. The severity of this failure is high because the frequency of the occurrences of these names is very high due to their popularity in Arabic names.

In order to improve the proposed framework, two steps could be done. They are reordering the iterations in relax condition and adding the first name in relax condition and then use Levenshtein method as a final step to get the best match name.

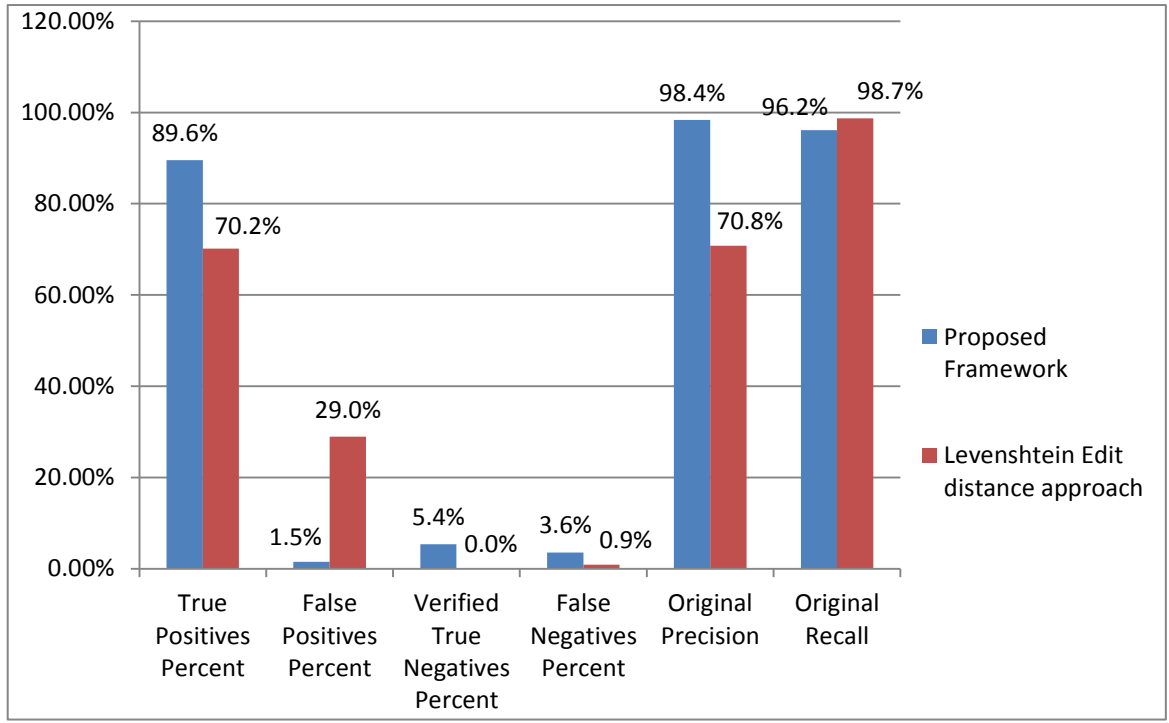

Figure 7: Results of comparing the proposed framework with Levenshtein Edit Distance Approach

*5.2.2. Experiment 2: Effect of dictionary building technique on the quality of Names Matching*

Several dictionary building approaches are used. The first one uses the source dataset to extract the dictionary. This approach suffer from the data entry errors, resulting in 40% for one to one true positives matching and 44.8% overall true positive accuracy percentage. The second approach uses also the source dataset and then applies the soundex algorithm to remove any records that do not have matched Soundex for both English and Arabic names. The results of using this dictionary was 45% for one to one true positives matching and 51% overall true positive accuracy percentage. These results show that the Soundex based dictionary has moderate performance. However, some records are not matched due to the complexity of the Arabic names with several syllables. The third dictionary uses Arabic combined Soundex algorithm to match Arabic and English transliterated Arabic Names. The results of using this dictionary was 70% for one to one true positives matching and 78% overall true positive accuracy percentage. The fourth dictionary uses the Arabic combined soundex algorithm and then a domain expert verifies the entries. The dictionary is used to match Arabic and English transliterated Arabic Names. The results of using this dictionary was 72% for one to one true positives matching and 80% overall true positive accuracy percentage.

| | Source Dataset Extracted Dictionary | Soundex Based Dictionary | Combined Soundex Based Dictionary | Domain Expert Verified Combined Soundex Dictionary |
|---|---|---|---|---|
| True Positives Percent | 40.55% | 44.62% | 69.86% | 72.01% |
| Overall True Positive Accuracy Percentage | 48.26% | 49.60% | 77.81% | 80.37% |

The obtained results show that data entry based dictionaries cannot be used lacks proof verification. Therefore, we recommend adding Soundex verification component in each data entry form that is used to enter person names with dual languages. Automatic dictionary building is a possible alternative in cross languages person names mapping although their results do not exceed dictionaries built and verified by human domain experts.

### 5.2.3. Experiment 3: English transliterated / Arabic Person Names Matching

The last experiment uses the digital library data that represents Egyptian scholars' data which is cited in Scopus and ISI web of knowledge. This data represent the **English transliterated names** and is matched with the **Arabic data found in UMIS**. The data is characterized with the existence of initial letters on most of records. For example, "A. M. Aly" instead of "Abdullah Mohamed Aly". Another example is "Hany, A. M.". The source dataset contains 1,000 records and the destination dataset has 80,000 records. The clerical review process enabled gathering of the expert opinions about matches, possible matches and non-matches. These opinions are then converted to metrics and compared to the metrics of machine to form the confusion matrix. The result of the confusion matrix of this experiment is shown below.

| | | | Machine | | | | | | | | | | |
|---|---|---|---|---|---|---|---|---|---|---|---|---|---|
| | | | M | U | P | | | | | | | | |
| | | | 1..1 | 1..0 | 1..2 | 1..3 | 1..4 | 1..5 | 1..6 | 1..7 | 1..8 | 1..9 | 1..10 |
| Subject Matter Expert | M | 1..1 | 615 | 46 | 38 | 15 | 4 | | | 1 | | | 1 |
| | U | 1..0 | 25 | 176 | 2 | 2 | 2 | 1 | 1 | 1 | 2 | | |
| | P | 1..2 | | | 16 | | | | | | | | |
| | | 1..3 | | | | 11 | | | | | | | |
| | | 1..4 | | | | | 3 | | | | | | |
| | | 1..5 | | | | | | | 2 | | | | |
| | | 1..6 | | | | | | | 1 | | | | |
| | | 1..7 | | | | | | | 3 | | | | |
| | | 1..8 | | | | | | | | | 2 | | |
| | | 1..9 | | | | | | | | | | 3 | |

The results show that the proposed framework was able to get about 62% true positives, 18% verified true negatives. The possible true positives (green row) are about 3%. The false positives were about 2.5% and false negatives were about 5%. These results show the feasibility, effectiveness and efficiency of the proposed framework to match about 90% of English transliterated names with middle initials to their corresponding Arabic Names. It is worth mentioning here that Levenshtein distance based algorithms cannot be used here because the distance between a name and its initial character will be very high, for example the distance between Mohamed and M is 6. So it will not be suitable for use in such problems.

| Metrics | Value |
|---|---|
| True Positives Percent | 62% |
| False Positives Percent | 3% |
| Verified True Negatives Percent | 18% |
| False Negatives Percent | 5% |
| Extended True Positive Accuracy Percentage | 3% |
| Original Precision | 96% |
| Original Recall | 93% |
| Overall True Positive Accuracy Percentage | 64% |

## 6. Conclusion

In this paper, a framework for cross languages name mapping between English and Arabic was proposed and implemented. A new version of Arabic Soundex is implemented and used to support building a baseline dictionary from existing data. The results of applying the new version of Arabic Soundex show its ability to differentiate well between different Arabic combined names. A new proximity similarity matching decision model is proposed (Weighted Atomic Token) that is found to solve the need of respecting the names order in Arabic Language. New accuracy and effectiveness measures are proposed that superseded the well know quality metrics and suits better the cross languages name mapping problem. The record linkage quality of several case studies between Arabic/Arabic data, English/Arabic Data, English Abbreviated/ Arabic data showed the effectiveness of the proposed framework compared to other methods.

**Acknowledgments**

I thank both the Information and Communication Technology Project (ICTP) and the Egyptian Universities Library (EUL) unit for providing us with the real data used for the name matching process. I acknowledge also the work of Ziad Turkey, Azza Higazy and Sherehan Shams for their role in implementing and verifying the framework. We thank our colleagues who supported us in reviewing this paper.